\documentclass[letterpaper,10pt,journal,twoside]{IEEEtran}
\usepackage{amsmath,amsfonts}
\usepackage{algorithmic}
\usepackage{array}
\usepackage{blindtext}
\usepackage{textcomp}
\usepackage{stfloats}
\usepackage{url}
\usepackage{verbatim}
\usepackage{graphicx}
\def\BibTeX{{\rm B\kern-.05em{\sc i\kern-.025em b}\kern-.08em
    T\kern-.1667em\lower.7ex\hbox{E}\kern-.125emX}}
\usepackage{balance}
\usepackage{graphics}
\usepackage{subfigure}
\usepackage{booktabs}
\usepackage{multirow}
\usepackage{cite}
\usepackage{tikz,xcolor,hyperref}
\usepackage{algorithm}
\usepackage{fontawesome}
\usepackage{makecell}
\usepackage{tabularx}

\definecolor{lime}{HTML}{A6CE39}
\DeclareRobustCommand{\orcidicon}{%
	\begin{tikzpicture}
	\draw[lime, fill=lime] (0,0) 
	circle [radius=0.16] 
	node[white] {{\fontfamily{qag}\selectfont \tiny ID}};
	\draw[white, fill=white] (-0.0625,0.095) 
	circle [radius=0.007];
	\end{tikzpicture}
	\hspace{-2mm}
}

\foreach \x in {A, ..., Z}{%
	\expandafter\xdef\csname orcid\x\endcsname{\noexpand\href{https://orcid.org/\csname orcidauthor\x\endcsname}{\noexpand\orcidicon}}
}

\begin{document}
\title{Fully Asynchronous Neuromorphic Perception for Mobile Robot Dodging with Loihi Chips}

\author{Junjie Jiang\orcidA{}, Delei Kong\orcidB{}, Chenming Hu\orcidC{}, Zheng Fang\orcidD{}, \emph{Member, IEEE}
\thanks{
This work was supported in part by National Natural Science Foundation of China under Grant 62073066 and Grant U20A20197, in part by the Fundamental Research Funds for the Central Universities under Grant N2226001, in part by the 111 Project under Grant B16009, and in part by the Intel Neuromorphic Research Community (INRC) under Grant RV2.137.Fang. \emph{(Corresponding author: Zheng Fang.)}

Junjie Jiang, Chenming Hu and Zheng Fang are with Faculty of Robot Science and Engineering, Northeastern University, Shenyang, China (e-mail: jianggalaxypursue@gmail.com, hucm6108@gmail.com, fangzheng@mail.neu.edu.cn), Delei Kong is with School of Robotics, Hunan University, Changsha, China (e-mail: kong.delei.neu@gmail.com).

Digital Object Identifier (DOI): see top of this page.
}}


\maketitle
\begin{abstract}
Sparse and asynchronous sensing and processing in natural organisms lead to ultra low-latency and energy-efficient perception.
Event cameras, known as neuromorphic vision sensors, are designed to mimic these characteristics.
However, fully utilizing the sparse and asynchronous event stream remains challenging.
Influenced by the mature algorithms of standard cameras, most existing event-based algorithms still rely on the "group of events" processing paradigm (e.g., event frames, 3D voxels) when handling event streams.
This paradigm encounters issues such as feature loss, event stacking, and high computational burden, which deviates from the intended purpose of event cameras.
To address these issues, we propose a fully asynchronous neuromorphic paradigm that integrates event cameras, spiking networks, and neuromorphic processors (Intel Loihi).
This paradigm can faithfully process each event asynchronously as it arrives, mimicking the spike-driven signal processing in biological brains.
We compare the proposed paradigm with the existing "group of events" processing paradigm in detail on the real mobile robot dodging task.

Experimental results show that our scheme exhibits better robustness than frame-based methods with different time windows and light conditions.
Additionally, the energy consumption per inference of our scheme on the embedded Loihi processor is only 4.30\% of that of the event spike tensor method on NVIDIA Jetson Orin NX with energy-saving mode, and 1.64\% of that of the event frame method on the same neuromorphic processor.
As far as we know, this is the first time that a fully asynchronous neuromorphic paradigm has been implemented for solving sequential tasks on real mobile robot.
\end{abstract}
\def\abstractname{Note to Practitioners}
\begin{abstract}
As a neuromorphic visual sensor, the event camera offers a novel approach to achieving robust, low-power perception in robotics, owing to its high temporal resolution, wide dynamic range, and low information redundancy. 
However, the sparse and asynchronous nature of event streams presents a significant challenge for data processing. 
The mainstream approach preprocesses event streams into various representations (e.g., event frames, 3D voxels) before performing subsequent operations, leading to issues such as feature loss, event stacking, and high computational burden. 
In this paper, we propose a fully neuromorphic system that leverages the asynchronous characteristics of event streams and spiking neural networks to enable asynchronous processing of events.
Our asynchronous processing of event streams enhances robustness across different time windows and lighting conditions while significantly reducing power consumption during inference.
Our fully asynchronous neuromorphic system has been validated on a real mobile robot and is expected to advance the robot perception system towards biological perception.
\end{abstract}


\section{Introduction}
\label{sec:introduction}

By imitating asynchronous and sparse sensing and processing in natural organisms, neuromorphic technology is expected to achieve robust, efficient, and ultra low-power perception\cite{paredes2024fully}.
Event cameras, known as neuromorphic visual sensors, only asynchronously trigger events at pixel points where the light change exceeds a contrast threshold\cite{gallego2020event}. 
They have been proved to have the advantages of high temporal resolution, high dynamic range (HDR), low information redundancy, and low power consumption in many tasks.
However, how to utilize sparse asynchronous event stream effectively and efficiently remains a challenge.
With the significant success of artificial neural networks (ANNs) in conventional vision, researchers have processed events into various representations to align with the mature deep network frameworks for conventional vision\cite{gehrig2023recurrent, rebecq2019high, rudnev2023eventnerf, jiang2024ev}.
These approaches have achieved impressive results in fields such as optical flow estimation\cite{wan2022learning,zhuang2024ev,gehrig2024dense}, depth estimation\cite{hidalgo2020learning,nam2022stereo,zhang2022discrete}, visual place recognition\cite{kong2022event,hou2023fe,yu2023brain,hu2024spike} and so on\cite{guan2023pl,chen2021novel, wu2021novel}.
Regarding event representations, early researchers compressed the event stream to specific moments, converting it into a representation similar to an image frame\cite{maqueda2018event, tian2022event, sanket2020evdodgenet}.
However, these frame-based representations lose most of the temporal information in the event stream and can cause issues such as feature loss or event stacking when the time window is too short or too long.
To improve the utilization of the temporal information in the event stream, subsequent researchers proposed voxel-based representations\cite{zhu2019unsupervised, gehrig2019end}.
These voxel-based representations outperformed the frame-based representations on a series of tasks\cite{gehrig2019end}.
However, due to the sparseness of events, these voxel-based representations introduce a large number of redundant data where there are no events, significantly increasing the computational burden.

\begin{figure*}[htbp]
\vspace{-0.5em}
\centering
\includegraphics[width=\textwidth]{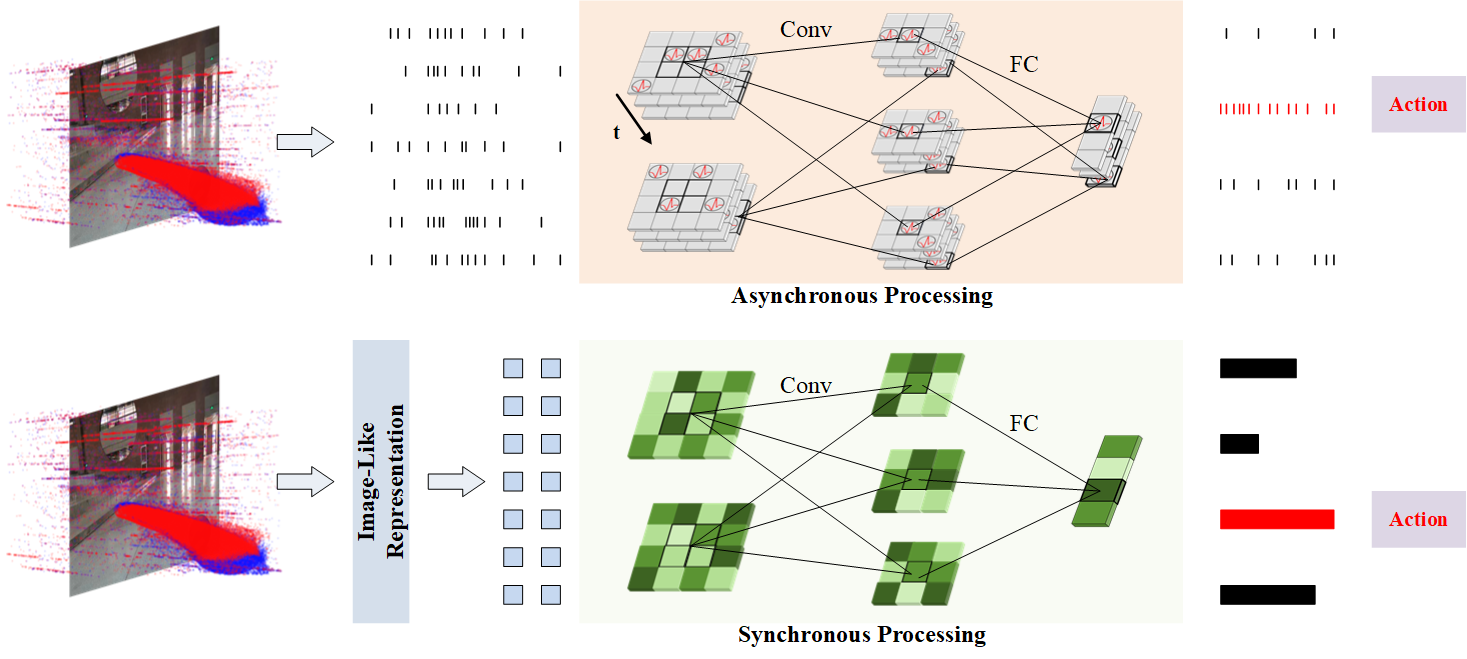}
\vspace{-2em}
\caption{Comparison between synchronous and asynchronous paradigms for event streams. 
In the asynchronous paradigm, events are processed asynchronously according to their triggered time. 
The triggered neurons send spikes that are processed asynchronously.
In the synchronous paradigm, event streams are converted to the image-like reprensentation and information is processed synchronously by every neuron.}
\label{fig:1}
\vspace{-0.5em}
\end{figure*}

Compared to ANNs, spiking neural networks (SNNs) inspired by neuromorphic computing are regarded as the next generation of neural network models due to their high biological plausibility, event-driven characteristics, and high energy efficiency\cite{roy2019towards, yao2023attention, yao2024spike, jiang2023neuro, ruan2024gsnn, yi2020tactile}.
Unlike ANNs that synchronously process floats among multi-layer neurons, SNNs process information asynchronously through spike signals, with each neuron exhibiting rich spatio-temporal dynamics.
When processing event streams, SNNs exhibit natural adaptability compared to ANNs.
They are expected to minimize delays and fully explore the spatial-temporal features of events.
In particular, with the rapid development of  Non-von Neumann architecture chips, a series of neuromorphic processors adhering to the principles of brain-like computing have been launched\cite{davies2018loihi, akopyan2015truenorth, chiavazza2023low}.
This advancement makes the on-chip sparse asynchronous processing of SNNs a reality.
Some recent works attempt to utilize SNNs on neuromorphic chips to process event stream\cite{viale2021carsnn, paredes2024fully}.
However, they do not fully combine the asynchronous characteristics of event streams and SNNs.
Specifically, \cite{viale2021carsnn,paredes2024fully} process events into frame-based representations and input the same representation into each time step of SNNs for processing.
The operation causes the sensitivity to time windows and significantly increases the computational burden of on-chip SNN, making it difficult to achieve effective and efficient neuromorphic visual perception.

To sum up, existing solutions struggle to utilize event streams effectively and efficiently.
The aforementioned issues pose significant challenges to small mobile robot systems, which have limited on-board computation resources and exhibit high temporal dependencies.
To address the above issues, we propose a fully asynchronous neuromorphic paradigm that integrates event cameras, spiking networks, and neuromorphic processors (Intel Loihi).
Our paradigm implements asynchronous processing of events with spiking networks according to their timestamps, as shown in Fig.~\ref{fig:1}.
Therefore, our paradigm utilizes temporal information in events without additional representations and minimizes redundant information to reduce the computational burden.
Our contributions are as follows:

\begin{itemize}
\item We propose an asynchronous fully neuromorphic paradigm and apply it to solve sequential tasks on the real mobile robot for the first time. Our paradigm integrates event cameras, spiking networks, and neuromorphic processors to utilize event streams efficiently and effectively. 
\item We propose a Key-Event-Point (KEP) module to extract the key event stream from the raw event stream using spatio-temporal clustering and Kullback-Leibler (KL) divergence.
Our module can retain the key information of the raw event stream while reducing the number of events, addressing the issue of limited transmission bandwidth when processing events on neuromorphic processors.
\item We propose a dodging network based on SNNs with offline training and online inference. In the offline training phase, we introduce a suitable loss function and employ fast backpropagation to address supervision and backpropagation challenges in SNNs. For online inference, we implement weight integer quantization and establish the mapping between events and neurons to enable on-chip asynchronous processing. The output spike trains are then decoded to determine the dodging direction and speed.

\item We conduct a detailed comparison of the comprehensive performance of various methods based on ANNs and SNNs with our method in the mobile robot dodging task. 
First, we evaluate the dodging success rate of different methods at different time windows, lighting conditions and objects. 
Then, we evaluate the computational power consumption of different methods across three devices: a desktop GPU (NVIDIA Geforce RTX 4090), an embedded processor (NVIDIA Jetson Orin NX), and a neuromorphic processor (Intel Kapoho Bay).
\end{itemize}

The rest of this article is organized as follows: Sec. \ref{sec:relatedwork} reviews the related work on various event stream processing approaches based on ANNs and SNNs. Sec. \ref{sec:methodology} describes the implementation of the fully asynchronous neuromorphic paradigm and the mobile robot system. Sec. \ref{sec:experiments} presents various experiments about mobile robot dodging to verify the advantages of the proposed paradigm. Finally, Sec. \ref{sec:conclusions} concludes the article.


\section{Related Work}
\label{sec:relatedwork}

The mainstream ANN-based framework for vision is synchronous, making it difficult to process asynchronous event stream. Therefore, some researchers converted event streams into various representations to make them compatible with ANNs.
For example, Tian et al. \cite{tian2022event} fixed the number of events to compress the event stream into the event frame, subsequently realizing optical flow estimation through a transformer network architecture. 
Sanket et al. \cite{sanket2020evdodgenet} averaged the events within a fixed time window to obtain the intensity value and used an additional channel to record the average timestamps of events for dynamic object perception. 
However, these frame-based methods lose the bionic characteristics of vision and are heavily dependent on scenes and motion, resulting in poor robustness.
Zhu et al. \cite{zhu2019unsupervised} proposed a voxel representation of the event stream to improve the utilization of temporal information. 
They mapped the event stream to corresponding voxels using a linear weighted accumulation method of bilinear interpolation, achieving better results in optical flow estimation and depth estimation compared to the event frame methods. 
Gehrig et al.\cite{gehrig2019end} introduced a learnable convolution kernel to convert the event stream into a voxel representation, e.g., event spike tensor (EST), achieving superior performance in both recognition and optical flow estimation tasks compared to event frame methods. 
However, these methods introduce an additional three-dimensional voxel grid, increasing the computational burden. 
Limited by the synchronous processing mechanism of ANNs, various event processing methods based on ANNs struggle to utilize event streams effectively while maintaining computational efficiency.

Compared to ANNs, SNNs exhibit high biological plausibility, event-driven, and high energy efficiency characteristics that are consistent with event streams. 
Researchers began to explore how to combine event stream and SNNs. 
Initially, researchers focused on converting pre-trained ANNs into SNNs. 
For example, Massa et al.\cite{massa2020efficient} proposed a scheme for converting pre-trained ANNs into SNNs, enabling event-based gesture recognition using SNNs. 
However, the performance of the converted SNN depends heavily on the fine-tuning of the spiking frequency of the input layer and the threshold of each layer of neurons\cite{roy2019towards}. 
With the introduction of direct training methods of SNNs\cite{wu2018spatio, shrestha2018slayer}, Cuadrado et al.\cite{cuadrado2023optical} achieved event-based optical flow with direct trained SNNs. 
Gehrig et al.\cite{gehrig2020event} successfully realized angular velocity regression of event streams based on directly trained SNNs using the SLAYER framework\cite{shrestha2018slayer}. 
However, the SNNs in these studies were all implemented using GPU simulation, which maked it difficult to show the asynchronous potential and energy-efficient on the actual system. 
With the continuous advancement of neuromorphic processors\cite{davies2018loihi, akopyan2015truenorth}, researchers further explored the fully neuromorphic system that integrates event cameras, spiking networks and neuromorphic processors. 
For instance, Viale et al.\cite{viale2021carsnn} used fully neuromorphic system to extract cars from the background.
Paredes-Vallés et al.\cite{paredes2024fully} built neuromorphic system to estimate the optical flow of corner points.
\cite{viale2021carsnn, paredes2024fully}did this by accumulating event streams within a fixed time window and inputting the corresponding event frames into SNNs at different time steps. 
However, the current practice of converting event streams into event frames and then feeding them into SNNs at different time had several drawbacks. 
It faced issues like feature loss or event staking due to inappropriate time windows. 
Additionally, this method introduces a significant computational burden as the time step of the SNN increases.

In contrast to previous work, we achieved asynchronous processing of event streams in on-chip SNNs integrating the event camera, the SNN and the Intel Loihi processor. 
Our asynchronous paradigm makes better use of event streams, showing better robustness at different time windows, lighting conditions and objects while significantly reducing the computational burden.

\section{Methodology}
\label{sec:methodology}

\begin{figure}[htbp]
\vspace{-0.5em}
\centering
\includegraphics[width=\columnwidth]{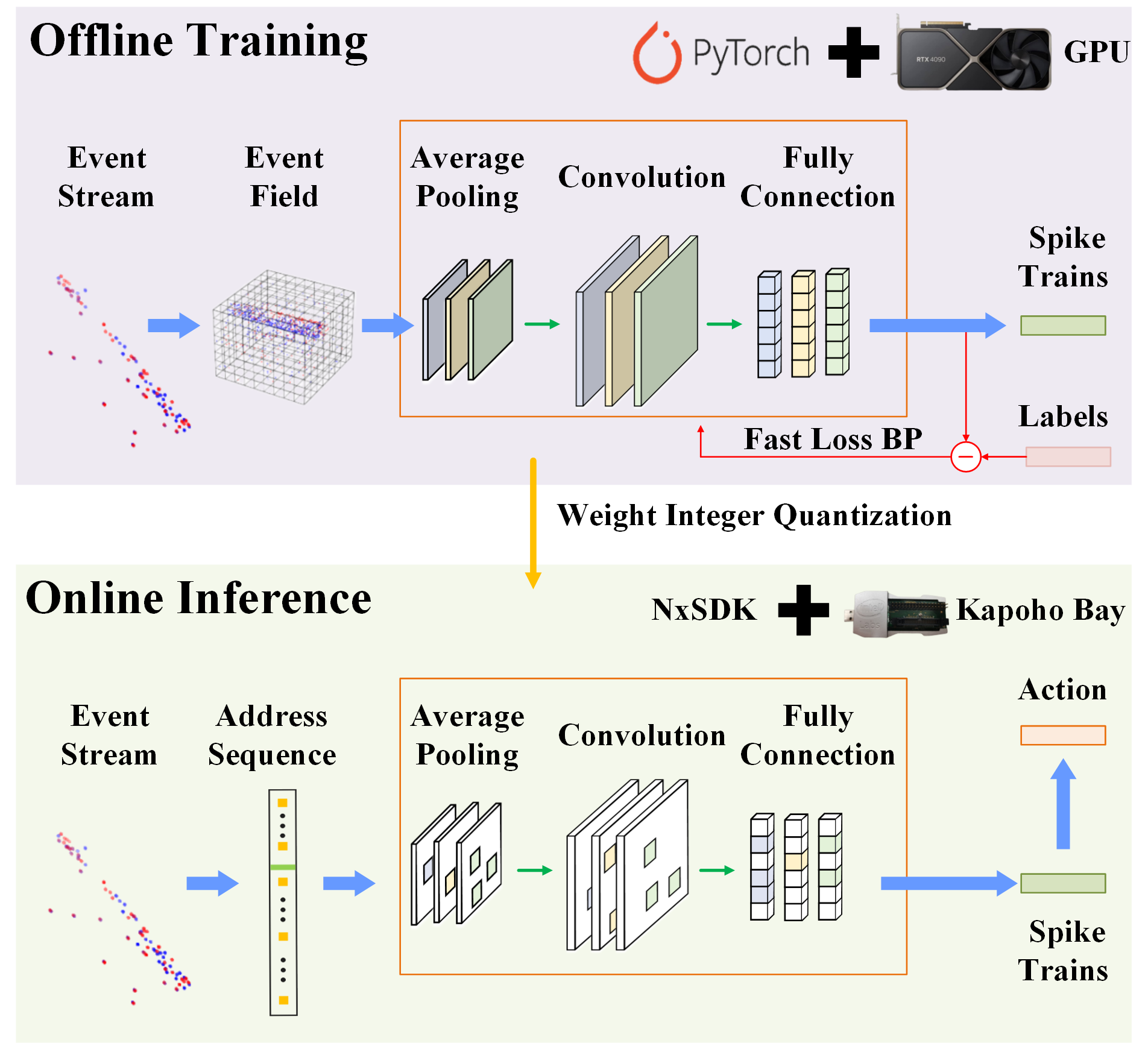}
\vspace{-2.0em}
\caption{Offline training and online inference of dodging network. 
The offline simulated SNN is trained on GPU with a specific loss function and fast loss backpropagation. 
Then the pre-trained SNN is deployed to Kapoho Bay using weight integer quantization to achieve online inference. 
Finally, the output spike trains are decoded to dodging action.}
\label{fig:2}
\vspace{-0.5em}
\end{figure}

In this section, we describe how to implement the fully asynchronous neuromorphic paradigm for achieving mobile robot dodging. 
First, we introduce the event camera and Loihi processor (Kapoho Bay).       
Then, we introduce the Key-Event-Point (KEP) module, which extracts key event streams from the raw event streams to reduce the number of events. 
After that, we present the details of the dodging network training and inference as shown in Fig.~\ref{fig:2}. 
Finally, we introduce the fully neuromorphic mobile robot system for dodging.

\subsection{Event Camera and Kapoho Bay}
\subsubsection{Event Camera}
Event cameras are neuromorphic visual sensors that mimic the biological retina and use the asynchronous data format called address event representation (AER) to simulate the transmission of neural signals in biological visual systems.
The pixel array of the event camera independently responds to pixel-level brightness changes in a logarithmic manner and triggers the sparse asynchronous event stream.
The brightness change of pixel $\left(x_{k}, y_{k}\right)^{\top}$ at time $t_{k}$ is given by:
\begin{equation}
\label{eq:1}
\begin{aligned}
\Delta \boldsymbol{L}\left(x_{k}, y_{k}, t_{k}\right)
&=\boldsymbol{L}\left(x_{k}, y_{k}, t_{k}\right) 
- \boldsymbol{L}\left(x_{k}, y_{k}, t_{k}-\Delta t_{k}\right),
\end{aligned}
\end{equation}
when the brightness change exceeds the threshold, the corresponding pixel triggers an event. 
The polarity of the event $p_{k}$ is determined by the brightness change as follows: 
\begin{equation}
\label{eq:2}
    p_{k}=
    \begin{cases}
    \text{ON},   &\Delta \boldsymbol{L}\left(x_{k}, y_{k}, t_{k}\right)>\vartheta\\
    \text{OFF},   &\Delta \boldsymbol{L}\left(x_{k}, y_{k}, t_{k}\right)<-\vartheta
    \end{cases},
\end{equation}
where $\vartheta$ is the contrast threshold. 
Each event point in the event stream is described by the pixel position, trigger time and polarity. 
The address event is represented as $\boldsymbol{e}_{k}=\left(x_{k}, y_ {k},t_{k},p_{k}\right)^{\top}.$

\subsubsection{Kapoho Bay}
Kapoho Bay is a portable neuromorphic processor developed by Intel, featuring two Loihi chips for 262144 neurons and 260 million synapses.
Loihi is a neuromorphic chip developed by Intel Labs\cite{davies2018loihi}.
It performs adaptive, self-modifying, event-driven computations with fine-grained parallelism, enabling efficient learning and inference. Each chip consists of 128 neuromorphic cores, fabricated using Intel's 14 nm process, and three embedded Lakemont (x86) processors used for I/O spike management.

\subsection{Key-Event-Point (KEP) Module}
To obtain key events that represent the current movement trend from the large number of raw event streams, we extract the event stream in two steps: main event stream extraction and key event stream extraction.

\subsubsection{Main Event Stream Extraction}
We remove events irrelevant to the object to extract the main event stream. 
Irrelevant events, such as noise points, are randomly distributed in the event stream. 
However, the motion event stream is densely distributed in a certain area of an event stream, thus we use the idea of clustering to determine the main event stream. 
Unlike simple planar clustering, the event stream contains the location of the triggering pixel $x_i, y_i$, the triggering time $t_i$ and the corresponding polarity $p_i$. 
We regard positive and negative polarities as equally important information, and the time dimension as the third dimension information, and perform spatio-temporal clustering on the event stream. 
We first calculate the optimal cluster center $\boldsymbol{u^{*}}=\left(x^{*}, y^{*}, t^{*}\right)$ of the event stream:
\begin{equation}
\label{eq:4}
\boldsymbol{u}^* = \underset{\boldsymbol{u}}{\arg\min} \sum_{i=1}^{N} \|\boldsymbol{e}\left(x_i,y_i,t_i\right) - \boldsymbol{u}\|^2,
\end{equation}
we get the $M$ points according to setting Euclidean distance to the cluster center as the main event stream.
\subsubsection{Key Event Stream Extraction}
To further speed up the inference, we extract the key event stream that can represent the movement trend from the main event stream. 
The number of events $M^{\prime}$ in the key event stream is determined by the following formula:
\begin{equation}
\label{eq:5}
    M^{\prime}=
    \begin{cases}
M     &\text{if } M<500 \\
\frac{1}{2}\lambda_{1}(1+e^{\frac{1}{M-\lambda_{1}}})   &\text{if } M< 1000 \text{ and }  M\geq 500\\
\frac{1}{2}\lambda_{2}(1+e^{\frac{1}{M-\lambda_{2}}})   &\text{if } M\geq 1000
\end{cases},
\end{equation}
where $\lambda_{1}=300$ and $\lambda_{2}=600$. We use KL divergence\cite{kullback1951information} to evaluate the similarity between the key and main event stream. 
To calculate the distribution probability of the event stream, we divide the three-dimensional space into $20\times 20\times 20$ cells. 
Then, we count the proportion of events in each cell relative to the total number of events to obtain the probability distribution. 
The probability distribution of the event stream is defined as follows:
\begin{equation}
\label{eq:6}
    p=\frac{h_i}{\sum_{i=1}^{K^3}{h_i}},
\end{equation}
where $h_i$ represents the number of events in the $i-th$ space. 
The KL divergence of the event stream is defined as follows:
\begin{equation}
\label{eq:7}
D_{KL}\left(p_{M^{\prime}}|p_M\right)=\sum_{x}p_{M^{\prime}}(x)\log\left(\frac{p_{M^{\prime}}(x)}{p_{M}(x)}\right),
\end{equation}
we randomly select $M^{\prime}$ several times in the main event stream and take the set of events with the lowest KL divergence as the key event stream.

\subsection{Dodging Network Offline Training}

Efficient and convenient supervised training on Kapoho Bay is challenging. 
Therefore, we simulate and train the SNN on a GPU during offline training. 
We complete offline training in the following parts: simulated SNN forward, SNN loss function and fast loss backpropagation.

\subsubsection{Simulated SNN Forward}
We use dual-state leaky-integrate-and-fire neurons\cite{maass1997networks, jiang2023neuro} to construct simulated SNN. 
The iterative update equation of the layer $n \in[1, \mathrm{N}]$ is determined by the following equation:
\begin{equation}
\label{eq:8}
\begin{aligned} 
\boldsymbol{C}_{n}^{t} &=\delta_{\text{curr}, n} \boldsymbol{C}_{n}^{t-1} + \boldsymbol{W}_{n} \boldsymbol{O}_{n-1}^{t}, \\
\boldsymbol{U}_{n}^{t} &=\delta_{\text{volt}, n} \boldsymbol{U}_{n}^{t-1} \boldsymbol{g}\left(\boldsymbol{O}_{n}^{t-1}\right)+\boldsymbol{C}_{n}^{t},    \\ 
\boldsymbol{O}_{n}^{t} &=\boldsymbol{h}\left(\boldsymbol{U}_{n}^{t}\right),
\end{aligned}
\end{equation}
where $t \in[1, \mathrm{T}]$ denotes discrete time, $\boldsymbol{O}_{1}^{t}=\boldsymbol{O}_{\text{in}, k}^{t}$ is the input spike trains, $\boldsymbol{O}_{\text{out},k}^{t}=\boldsymbol{O}_\mathrm{N}^{t}$ is the output spike trains. 
In order to achieve asynchronous processing of event streams, the time step $\mathrm{T}$ of the SNN needs to match the time window $\mathrm{W}$ of the event stream.
A spike train is a combinatorial sequence of spikes and silences. 
$\boldsymbol{W}_n$ is the synaptic weight matrix, $\boldsymbol{C}_{n}^{t}$ and $\boldsymbol{U}_{n}^{t}$ represent the membrane current and voltage. 
$\delta_{\text{curr}, n} $ and $\delta_{\text{volt}, n}$ are the decay coefficients of membrane current and voltage respectively. $\boldsymbol{g}\left(\boldsymbol{O}_{n}^{t-1}\right)=\mathbf{1}-\boldsymbol{O}_{n}^{t-1}$ is the reset gate. 
$\boldsymbol{h}\left(\boldsymbol{U}_{n}^{t}\right)=\boldsymbol{H}\left(\boldsymbol{U}_{n}^{t}-u_{\text{th}, n}\right)$ is the fire gate.
$u_{\text{th}, n}$ is the spike-triggered threshold.
This iterative update equation incorporates all behaviors (integration, firing, decay, and reset) of TS-LIF neurons. 
It can be seen that different from the activation functions of analog neurons such as ReLU, TS-LIF spiking neurons have obvious time dependencies. 

The simulated SNN based on GPU cannot realize asynchronous processing of event streams. 
Hence, the event stream is converted to the event field\cite{gehrig2019end} based on the triggered position and time, with two channels representing polarities. 
Each slice of the event field at time $t$ is used as the simulated SNN input $\boldsymbol{O}_{\text{in}, k}^{t}$.

\subsubsection{Loss Function}
We define the loss function as the difference between the actual and desired number of spikes within a time step $\mathrm{T}$. 
Specifically, we set the network's output channels $\mathrm{M}$ according to all possible approach directions of the dynamic object. 
The desired number of spikes for the output channel corresponding to the correct approach direction is set to $N_{\text{DT}} (0<N_{\mathrm{DT}}<\mathrm{T})$, and the desired number of spikes for the other channels is set to $N_{\text{DF}} (0<N_{\mathrm{DF}}<N_{\mathrm{DT}})$. 
Assuming that within the time step $\mathrm{T}$, the number of spikes obtained by each channel is $S=(N_{1},N_{2},\cdots,N_{L})$, and the desired number of spikes for each channel is $D=(N_{\mathrm{DF}},N_{\mathrm{DF}},\cdots,N_{\mathrm{DT}},\cdots,N_{\mathrm{DF}})$, then the loss $L$ is as follows:
\begin{equation}
\label{eq:9}
L=\dfrac{1}{2}\sum_{i=1}^\mathrm{M}\left(\dfrac{D[i]-S[i]}{\mathrm{T}}\right)^{2}.
\end{equation}

\subsubsection{Fast Loss Backpropagation}
To achieve fast loss backpropagation of the spiking neural network, we refer to \cite{shrestha2018slayer} to ignore the impact of the reset operation on the backpropagation. 
The impact of the spike input trains at any moment on the subsequent membrane potential can be obtained by the response convolution kernel based on the voltage and current decay coeﬀicients. 
First, we construct the discrete current response convolution kernel $\boldsymbol{\epsilon}_{\text{curr}}$ and the voltage response convolution kernel $\boldsymbol{\epsilon}_{\text{volt}}$ in a time step $\mathrm{T}$ according to the current decay coefficient $\boldsymbol{\epsilon}_{\text{curr}}$ and the voltage decay coefficient $\boldsymbol{\epsilon}_{\text{volt}}$:
\begin{equation}
\label{eq:10}
\begin{aligned}
    \boldsymbol{\epsilon}_{\text{curr}}[t] &= \delta_{\text{curr}}\boldsymbol{\epsilon}_{\text{curr}}[t-1], \\
    \boldsymbol{\epsilon}_{\text{volt}}[t] &= \delta_{\text{volt}}\boldsymbol{\epsilon}_{\text{volt}}[t-1]+\boldsymbol{\epsilon}_{\text{curr}}[t],
\end{aligned}
\end{equation}
where $\boldsymbol{\epsilon}_{\text{curr}}[0] = \boldsymbol{\epsilon}_{\text{volt}}[0]=1$. 
When considering the impact on backpropagation, the forward propagation equation can be rewritten as:
\begin{equation}
\label{eq:11}
\begin{aligned}
\boldsymbol{a}^{n}[t]&=\boldsymbol{W}^{n-1}\boldsymbol{s}^{n-1}[t],\\
\boldsymbol{u}^{n}[t]&=(\boldsymbol{\epsilon}_{\text{volt}}*\boldsymbol{a}^{n})[t],\\
\boldsymbol{s}^{n}[t]&=f_s(\boldsymbol{u}^{n}[t]),
\end{aligned}
\end{equation}
according to the rewritten forward propagation equation, the derivative of the loss to the weight is:
\begin{equation}
\label{eq:12}
\begin{aligned}
\dfrac{\partial{L}}{\partial{\boldsymbol{W}^{l}}}
&=\sum_{n=1}^{\text{T}}
\dfrac{\partial{L}}{\partial{\boldsymbol{a}^{l+1}}[n]}
\dfrac{\partial{a}^{l+1}[n]}{\partial{\boldsymbol{W}^{l}}}\\
&=\sum_{n=1}^{\text{T}}
\dfrac{\partial{L}}{\partial{\boldsymbol{a}^{l+1}}[n]}
{\boldsymbol{s}^{l}[n]}^{\top},
\end{aligned}
\end{equation}
we define the derivative of the loss $L$ to the weighted input spikes $\boldsymbol{a}^{l}[n]$ as: 
\begin{equation}
\label{eq:13}
\boldsymbol{d}^{l}[n]=\dfrac{\partial{\boldsymbol{L}}}{\partial{\boldsymbol{a}^{l}[n]}},
\end{equation}
because the weighted input spikes $\boldsymbol{a}^{l}[n]$ and membrane potential $\boldsymbol{u}^{l}[n]$ only affect the subsequent moments $t=n,n+1,\cdots,\text{T}$, thus $\boldsymbol{d}^{l}[n]$ can be expanded as:
\begin{equation}
\label{eq:14}
\boldsymbol{d}^{l}[n]=
\sum_{m=n}^\mathrm{T}
\dfrac{\partial{\boldsymbol{L}}}{\partial{\boldsymbol{s}^{l}[m]}}
\dfrac{\partial{\boldsymbol{s}^{l}[m]}}{\partial{\boldsymbol{u}^{l}[m]}}
\dfrac{\partial{\boldsymbol{u}^{l}[m]}}{\partial{\boldsymbol{a}^{l}[n]}},
\end{equation}
it is worth noting that the spikes between $\boldsymbol{u}$ and $\boldsymbol{s}$ are generated by the spike generation function $f_s$, which only considers the membrane potential of the neuron at the current moment: 
\begin{equation}
\label{eq:15}
\frac{\partial{\boldsymbol{s}}^{l}[m]}{\partial{\boldsymbol{u}}^{l}[m]}=f_s^{\prime}(\boldsymbol{u}^{l}[m]),
\end{equation}
we define $\dfrac{\partial{\boldsymbol{L}}}{\partial{\boldsymbol{s}^{l}[n]}}$ as $\boldsymbol{e}^{l}[n]$, and further expand $\boldsymbol{e}^{l}[n]$ to obtain:
\begin{equation}
\label{eq:16}
\boldsymbol{e}^{l}[n]=\sum_{m=n}^{T}\frac{\partial{L}}{\partial{\boldsymbol{a}}^{l+1}[m]}\frac{\partial{a}^{l+1}[m]}{\partial{\boldsymbol{s}}^{l}[n]}=\boldsymbol{d}^{(l+1)}[n]
\boldsymbol{W}^{(l)},
\end{equation}
and $\dfrac{\partial{\boldsymbol{u}^{l}[m]}}{\partial{\boldsymbol{a}^{l}[n]}}$ can be further simplified to:
\begin{equation}
\label{eq:17}
\frac{\partial{\boldsymbol{u}^{l}[m]}}{\partial{\boldsymbol{a}^{l}[n]}}=\frac{\partial\sum_{k=1}^{m}{\boldsymbol{\epsilon}}[m-k]\boldsymbol{a}_{l}[k]}{\partial{\boldsymbol{a}^{l}[n]}}=\boldsymbol{\epsilon}[m-n],
\end{equation}
Eq. (\ref{eq:13}) can be finally written as
\begin{equation}
\label{eq:18}
\boldsymbol{d}^{l}[n]=\sum_{m=n}^{T} \boldsymbol{e}^{l}[m]f_s^{\prime}(\boldsymbol{u}^{l}[m])\boldsymbol{\epsilon}[m-n],
\end{equation}
we use the spike escape function as the approximate alternative gradient function of the spike generation function $f_s$, that is
\begin{equation}
\label{eq:19}
    f_s^{\prime}(\boldsymbol{u}[m])=\frac{\tau_{n}}{\tau_{d}u_{\text{th}}}e^{-\frac{\left|\boldsymbol{u}[m]-u_{\text{th}}\right|}{\tau_{d}u_{\text{th}}}},
\end{equation}
where $\tau_{n}$ is the neuron time constant, $\tau_{d}$ is the derivative time constant. According to Eq. (\ref{eq:12}), (\ref{eq:18}), (\ref{eq:19}), the weight $\boldsymbol{W}^{l}$ can be updated.

\begin{figure}[htbp]
\vspace{-0.5em}
\centering
\includegraphics[width=0.5\textwidth]{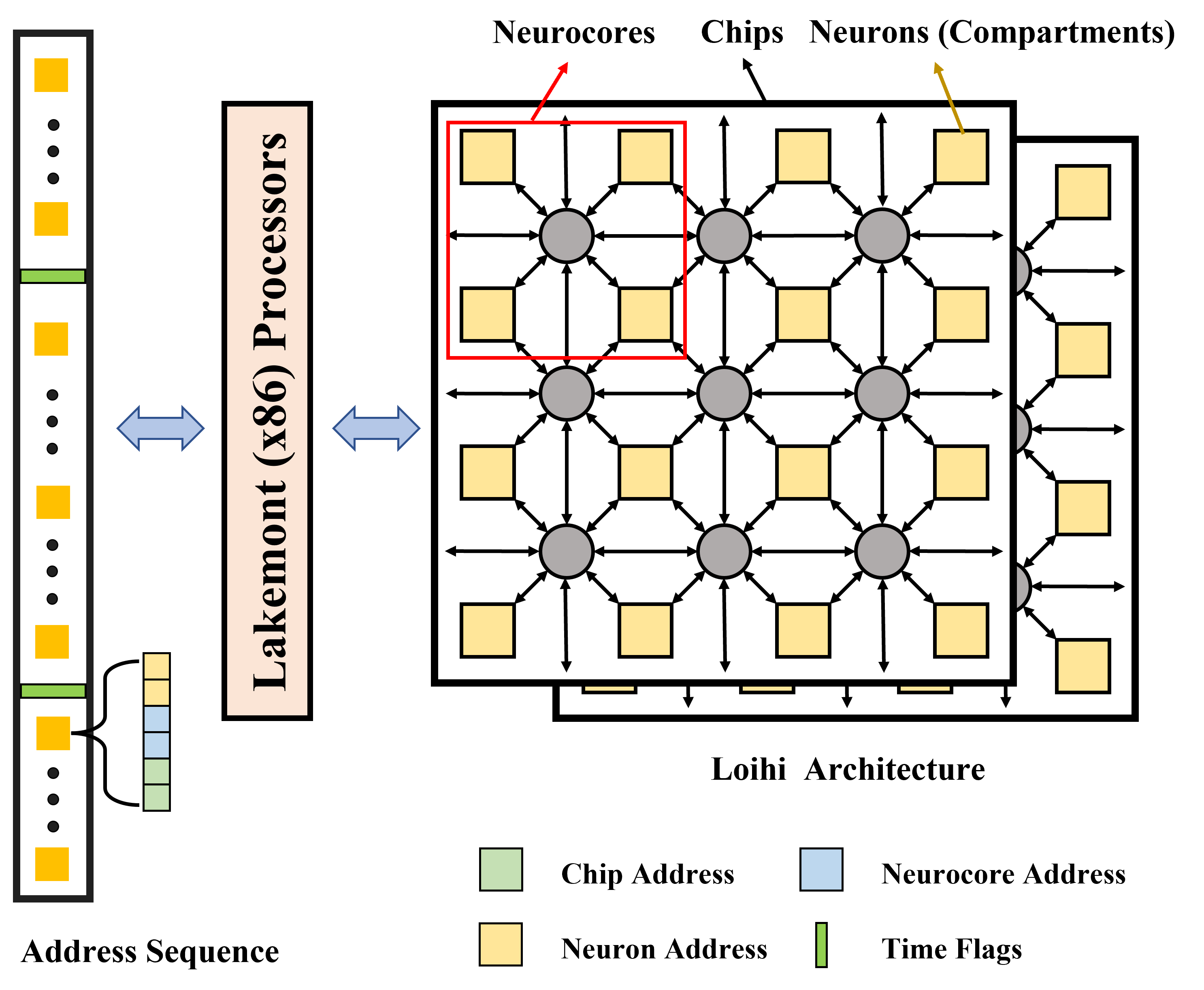}
\vspace{-2.0em}
\caption{Event camera asynchronous processing. 
The events first are converted to address sequence, containing the chip, neurocore, and neuron address. 
And time flags separate addresses at different times.
Finally, the event stream is sent to Loihi asynchronous with embedded lakemont (x86) processors.}
\label{fig:3}
\vspace{-0.5em}
\end{figure}

\begin{figure*}[htbp]
\vspace{-0.5em}
\centering
\includegraphics[width=\textwidth]{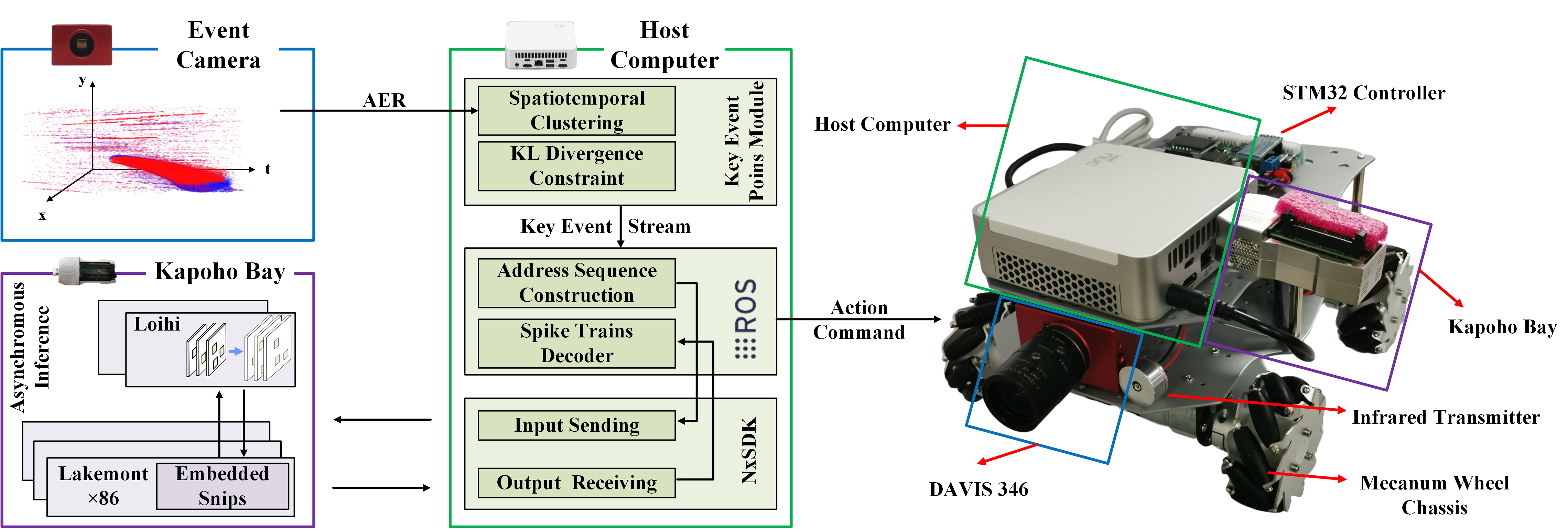}
\vspace{-2.0em}
\caption{The fully neuromorphic mobile robot system consisting of a mecanum wheel chassis, a STM32 controller, a infrared transmitter, a host computer, an event camera (DAVIS 346), and a neuromorphic processor (Kapoho Bay).
The host computer extracts the key event stream from the raw event stream and converts the event stream into the address sequence. 
Then the host computer sends events to corresponding neurons with embedded x86 processors under NxSDK.
After Kapoho Bay completes the inference, the host computer decodes the dodging action based on the output spike trains.}
\label{fig:4}
\vspace{-0.5em}
\end{figure*}

\subsection{Dodging Network Online Inference}
In this section, we first deploy the simulated SNN to Kapoho Bay using integer quantization. Then, we establish the mapping relatonship between events and neurons to achieve asynchrounous processing. Finally, we map the spike output trains to the dodging direction and speed.


\subsubsection{Weights Integer Quantization}
The Kapoho Bay supports 8-bit integer weights.
Hence, the network weights trained in the Pytorch framework need integer quantization when deploying on Kapoho Bay. 
Integer quantization is the conversion of floating-point weights to integer weights. 
The process of integer quantization is as follows:
\begin{equation}
\label{eq:20}
    \boldsymbol{W}_{\text{q}}=\text{round}\left(\frac{\boldsymbol{W}}{\sigma}\right)\sigma,
\end{equation}
where $\sigma=2$ is the minimum interval of integer quantization. 
$\boldsymbol{W}$ is the original weight, and $\boldsymbol{W}_{\text{q}}$ is the weight after integer quantization. 

\subsubsection{Event Stream Asynchronous Processing}
To achieve asynchronous processing of event streams, we need to establish the mapping relationship between events and neurons. 
An event triggers at time $t$ is represented as $e(x_i,y_i,p_i,t)$. 
The events that trigger at the same time $t$ are converted to mapped address $A_i$ as follows:
\begin{equation}
\label{eq:3}
    A_i=2\times x_i\times l_{\text{H}}+2\times y_i+p_i,
\end{equation}
where $l_{\text{H}}$ is the width of the event stream resolution. 
The processor establishes a mapping relationship between the mapped address and the actual address and assigns actual chip, neural core and neuron addresses to each pixel point. 
The event stream is converted into an address sequence, where events at different timestamps are distinguished by a time flag. 

The embedded x86 processors then asynchronously launch spikes to the corresponding chip, neural core, and neuron at different times based on the address sequence as shown in Fig.~\ref{fig:3}.

\subsubsection{Action Mapping}
The network outputs spike trains with $L$ channels.
The channel with the largest number of spikes represents the object's approach direction. 
We choose the opposite direction of the approach direction as the dodge direction. 
When the dynamic object approaches the camera faster, it generates a denser event stream. 
In this case, a faster dodging speed is required.
Therefore, the dodging speed is determined by the number of spikes in the direct channel. 
The dodge speed is calculated using the following equation:
\begin{equation}
\label{eq:21}
    v_{\text{dodge}} = \alpha \frac{N_r}{N_{\text{DT}}},
\end{equation}
where $\alpha = 2$ is the speed factor, $N_r$ is the number of spikes in the direct channel.


\subsection{The Fully Neuromorphic Mobile Robot System}
\label{neu-robot}

The fully neuromorphic mobile robot is shown in Fig.~\ref{fig:4}, which consists of a Mecanum wheel chassis, a STM32 controller, an infrared transmitter, a host computer, an event camera (DAVIS 346), and a neuromorphic processor (Kapoho Bay). 
The host computer first uses spatio-temporal clustering and KL divergence constraints to extract the key event stream from the raw event stream and convert the event stream into the address sequence. 
Then, the host computer sends the address sequence to the embedded x86 processors to realize the asynchronous processing of the event stream under NxSDK. 
After the Kapoho Bay completes the asynchronous inference, it transmits the spike trains back to the host computer. 
The host computer determines the approaching direction of the dynamic object based on the output spike trains and gives a control command to the STM32, which finally controls the chassis to complete the dodging action.

\section{Experiments}
\label{sec:experiments}
In this section, we outline the comprehensive performance of our proposed paradigm using quantitative and qualitative results. 
Firstly, we detail the experimental setup including Dataset, Parameters Setting and Baseline Schemes. 
Then, we compare dodging sucess rates and inference power of our paradigm with other schemes. 
Futhermore, we conducted real-time mobile robots dodging in real world.

\subsection{Experimental Setup}

\subsubsection{Dataset}
We used the DVS ROS driver\footnote{DVS ROS Driver Website: \url{https://github.com/uzh-rpg/rpg_dvs_ros/}.} provided by Scaramuzza et al. and recorded balls and humans approaching the event camera from the left and right sides under different light conditions as the dodging dataset. We set four lighting conditions: indoor normal light (Indoor NL, 1200 Lux), indoor low light (Indoor LL, 15 Lux), outdoor normal light (Outdoor NL, 10000 Lux) and outdoor low light (Outdoor LL, 25 Lux). 
The visualization examples are shown in Fig.~\ref{fig:5}. Specifically, the event stream in the low light condition shows more noise than that in a normal environment. 
Among them, we only use the ball dataset under normal lighting conditions for training, and test the dodging success rates of the ball and people under different light conditions respectively. 
Our training datasets and test datasets are shown in Table~\ref{Table:1}.
When recording the dataset, we adjusted the resolution of the raw event stream to $128\times128$ and used two channels to represent the positive and negative polarity of the event stream. 
\subsubsection{Parameters Setting}
We use dual-state leaky-integrate-and-fire neurons to construct the dodging network for training and inference.
The specific spiking neural network architecture is shown in Table~\ref{Table:2}.
The parameters of spiking neurons and training network are shown in Table~\ref{Table:3}.
\subsubsection{Baseline Schemes}
We set up comparison schemes from representations and networks. Specifically, we choose the event frame \cite{viale2021carsnn} and event spike tensor \cite{gehrig2019end} as representations. We select the event frame method based on the ANN, the event spike tensor method based on the ANN and event frame method based on the SNN as comparison schemes. 
In the following discussion, we refer to the event-frame method and event spike tensor method based on ANNs as EF-ANN and EST-ANN, respectively. The event-frame method based on SNNs is referred to as EF-SNN.

\begin{figure*}[htbp]
\vspace{-0em}
\centering
\subfigure[Image Frame]{\label{fig:5a}{\includegraphics[width=0.15\textwidth]{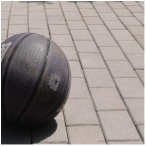}}}
\subfigure[Raw Event Stream]{\label{fig:5b}\frame{{\includegraphics[width=0.15\textwidth]{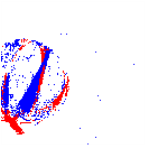}}}}
\subfigure[Key Event Stream]{\label{fig:5c}\frame{{\includegraphics[width=0.15\textwidth]{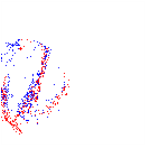}}}}
\subfigure[Image Frame]{\label{fig:5d}{\includegraphics[width=0.15\textwidth]{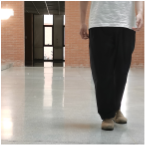}}}
\subfigure[Raw Event Stream]{\label{fig:5e}\frame{{\includegraphics[width=0.15\textwidth]{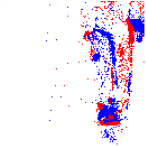}}}}
\subfigure[Key Event Stream]{\label{fig:5f}\frame{{\includegraphics[width=0.15\textwidth]{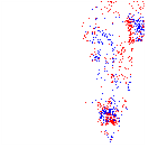}}}}
\subfigure[Image Frame]{\label{fig:5g}{\includegraphics[width=0.15\textwidth]{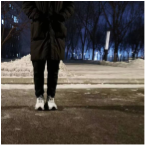}}}
\subfigure[Raw Event Stream]{\label{fig:5h}{\frame{\includegraphics[width=0.15\textwidth]{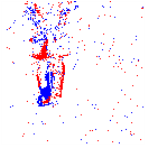}}}}
\subfigure[Key Event Stream]{\label{fig:5i}\frame{\includegraphics[width=0.15\textwidth]{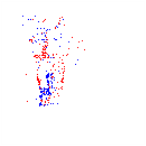}}}
\subfigure[Image Frame]{\label{fig:5j}{\includegraphics[width=0.15\textwidth]{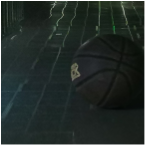}}}
\subfigure[Raw Event Stream]{\label{fig:5k}\frame{{\includegraphics[width=0.15\textwidth]{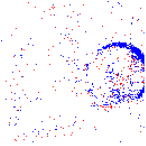}}}}
\subfigure[Key Event Stream]{\label{fig:5l}\frame{{\includegraphics[width=0.15\textwidth]{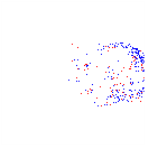}}}}
\caption{Some frame images, event frames of the raw and key event stream of datasets under outdoor normal light (10000 Lux), indoor normal light (1200 Lux), outdoor low light (25 Lux) and indoor low light (15 Lux). Frame images are only for visualization. Raw and key event streams are converted to event frames for visualization only. (a)-(c) are under outdoor normal light. (d)-(f) are under indoor normal light. (g)-(i) are under outdoor low light. (j)-(l) are under indoor low light.}
\label{fig:5}
\vspace{-1em}
\end{figure*}

\begin{table}[htbp]
\vspace{-1.0em}
\begin{center}
\caption{
Dodging Datasets.
}
\vspace{-1.0em}
\setlength{\extrarowheight}{3pt}
\setlength{\tabcolsep}{0.05\linewidth}
\begin{tabular}{ccccp{1cm}}
\Xhline{1pt}
\textbf{Objects}
&\textbf{Scenes}
&\textbf{Set}
&\textbf{Usuage}\\
\Xhline{0.7pt}
Ball    &normal light / indoor   &800  &training\\
Ball    &normal light / indoor   &800    &evaluation\\
Ball    &low light / indoor       &800    &evaluation\\
Ball    &normal light / outdoor   &800    &evaluation\\
Ball    &low light / outdoor      &800    &evaluation\\
Human   &normal light / indoor    &800    &evaluation\\
Human   &low light / indoor       &800    &evaluation\\
Human   &normal light / outdoor   &800    &evaluation\\
Human   &low light / outdoor      &800    &evaluation\\
\Xhline{1pt}
\end{tabular}
\label{Table:1}
\end{center}
\end{table}


\begin{table}[htbp]
\vspace{-1.0em}
\begin{center}
\caption{
SNN Architecture.
}
\vspace{-1.0em}
\setlength{\extrarowheight}{3pt}
\setlength{\tabcolsep}{0.03\linewidth}
\begin{tabular}{ccccccp{1cm}}
\Xhline{1pt}
\textbf{\makecell[c]{Layer\\Type}}&\textbf{\makecell[c]{Input\\Channel}} &\textbf{\makecell[c]{Output\\Channel}}&\textbf{\makecell[c]{Kernel\\Size}} & \textbf{Padding} &\textbf{Stride}\\
\Xhline{0.7pt}
AvgP@T    &2    &2    &4  &\textemdash &\textemdash\\
Conv@T    &2    &16   &3  &1           &1\\
AvgP@T    &16   &16   &2  &\textemdash &\textemdash\\
Conv@T    &16   &32   &3  &1           &1\\
AvgP@T    &32   &32   &2  &\textemdash &\textemdash\\
FC@T      &2048 &512  &\textemdash &\textemdash &\textemdash\\
FC@T      &512  &2    &\textemdash &\textemdash &\textemdash\\
\Xhline{1pt}
\end{tabular}
\label{Table:2}
\end{center}
\end{table}

\begin{table}[htbp]
\vspace{-1.0em}
\begin{center}
\caption{
Training Hyper-parameters.
}
\vspace{-1.0em}
\setlength{\extrarowheight}{3pt}
\setlength{\tabcolsep}{0.06\linewidth}
\begin{tabular}{ccccccp{1cm}}
\Xhline{1pt}
\textbf{Hyper-Parameters}  &\textbf{Values} \\
\Xhline{0.7pt}
Membrane current decay coefficient $\delta_{\text{curr}}$ & 0.75 \\
Membrane voltage decay coefficient $\delta_{\text{volt}}$ & 0.96875 \\
Spike-triggered threshold $u_{\text{th},n}$ & 0.8 \\
Desired spikes number for direct channel $N_{\text{DT}}$ & 25, 30, 70 \\
Desired spikes number for other channel $N_{\text{DF}}$ & 5, 10, 10 \\
Learning rate $\text{lr}$ & 0.001 \\
Neuron time constant $\tau_{n}$ & 1 \\
Derivative time constant $\tau_{d}$ & 1.25 \\
\Xhline{1pt}
\end{tabular}
\label{Table:3}
\end{center}
\end{table}

\subsection{Dodging Experiments}
In dodging experiments, we first compare the dodging success rates with different time windows, lighting conditions and objects. 
We then show the visualization effect after applying our Key-Event-Stream (KEP) module, as well as the changes in success rates and the number of events. 
Finally, we compare power consumption across three different devices: 1) NVIDIA GeForce RTX 4090 GPU, 2) NVIDIA Jetson Orin NX, and 3) Intel Kapoho Bay neuromorphic processor. In \ref{sec:dodge experiment} and \ref{sec:kep module}, EF-ANN and EST-ANN are implemented on  NVIDIA GeForce RTX 4090 GPU, while EF-SNN and our proposed paradigm are implemented on Intel Kapoho Bay neuromorphic processor. It should be noted that our KEP module effectively reduces the number of events while removing noise. 
To better verify the use of temporal information by different schemes, we evaluate the dodging success rates without using the KEP module in \ref{sec:dodge experiment}.

\subsubsection{Dodging Success Rates}
\label{sec:dodge experiment}

\begin{figure}[htbp]
\vspace{-0.5em}
\centering
\includegraphics[width=\columnwidth]{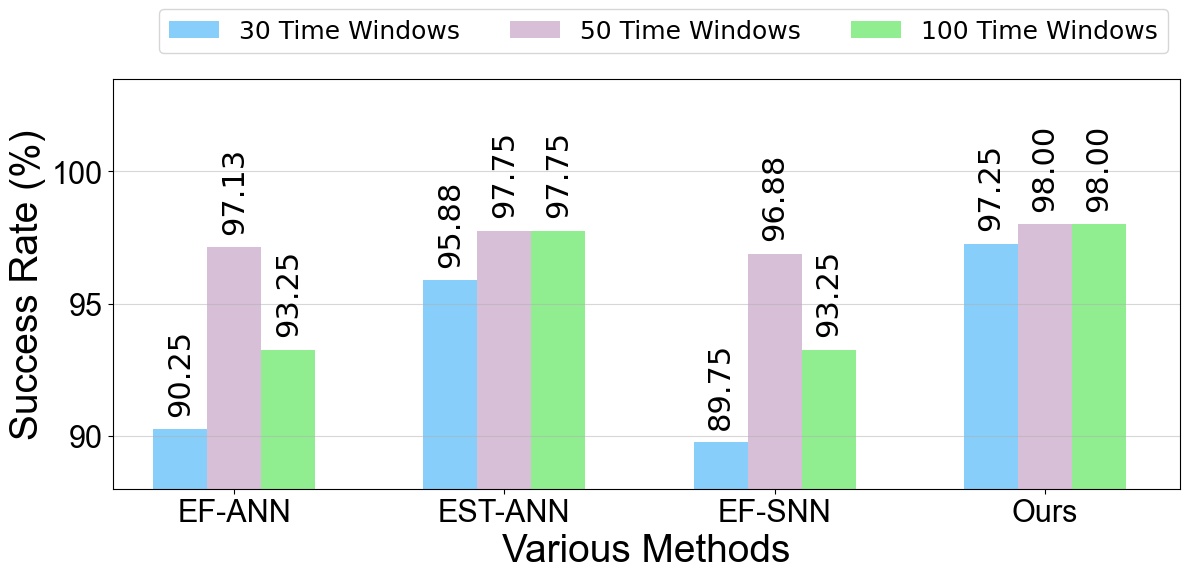}
\vspace{-2.0em}
\caption{Dodging success rate of different schemes with different time windows.}
\label{fig:6}
\vspace{-0.5em}
\end{figure}

\begin{figure}[htbp]
\vspace{-0mm}
\centering

\subfigure[Failure example at $\mathrm{T}=30$]{\label{fig:7a}\frame{{\includegraphics[width=0.23\textwidth]{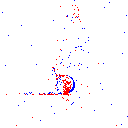}}}}
\subfigure[Failure example at $\mathrm{T}=100$]{\label{fig:7b}\frame{{\includegraphics[width=0.23\textwidth]{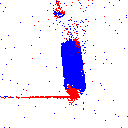}}}}

\vspace{0em}
\caption{Ball dodging failure cases based on EF-ANN at $\mathrm{T}=30$ and $\mathrm{T}=100$.}
\label{fig:7}
\vspace{0em}
\end{figure}

In dodging success rates experiments, we evaluate dodging success rates with different time windows, lighting conditions and objects.
First, we compared the performance of EF-ANN, EST-ANN EF-SNN and our method with different time windows (30, 50, 100) under indoor normal lighting. The results are shown in Fig.~\ref{fig:6}.
EF-ANN performs well only at the time window $\mathrm{W}=50$, with the success rate dropping at the time window $\mathrm{W}=30$ and $\mathrm{W}=100$. 
EST-ANN performs best at the time window $\mathrm{W}=50$ and $\mathrm{W}=100$, but the success rate drops by 1.87\% at the time window $\mathrm{W}=30$. 
EF-SNN performs similarly to EF-ANN. 
Our method achieves a success rate of 97.25\% at time window $\mathrm{W}=30$, and a success rate of 98.00\% at time window $\mathrm{W}=50$ and $\mathrm{W}=100$. 
When the time window is $\mathrm{W}=30$, the success rates of EF-ANN and EF-SNN drop significantly because, at slow ball speeds, the event frames formed in a short time interval make it difficult to judge the motion trend, as shown in Fig.~\ref{fig:7a}. 
When the time window is $\mathrm{W}=100$, the success rates of EF-ANN and EF-SNN also drop significantly because, with fast ball speeds, a longer time window leads to event stacking, as shown in Fig.~\ref{fig:7b}. 
EST-ANN and our method show good robustness with different time windows. 

We then selected the best model of each scheme to evaluated dodging success rates with different lighting conditions and objects. 
The results are shown in Table~\ref{Table:4}.
Experimental results show that EF-ANN and EF-SNN both show a decrease in success rate in indoor low-light scenes for the ball. 
This is because the indoor low light, around 15 lux, results in a very sparse event stream for the ball, making it highly susceptible to noise. 
Notably, in this condition, the ball-dodging success rate of EF-SNN decreases by 6.93\%, while EF-ANN only sees a 3.5\% reduction compared to the normal light condition.
This further demonstrates the incompatibility between the frame-based methods and SNNs.

\begin{table}[htbp]
\vspace{-1.0em}
\begin{center}
\caption{
Dodging Success Rates of Different Schemes with Different Lighting Conditions and Objects.
}
\vspace{-1.0em}
\setlength{\extrarowheight}{2pt}
\setlength{\tabcolsep}{0.03\linewidth}
\begin{tabular}{ccccccp{1cm}}
\Xhline{1pt}
\multirow{2}{*}{\textbf{Scheme}}
&\multirow{2}{*}{\textbf{Object}}
&\multicolumn{2}{c}{\textbf{Indoor}}
&\multicolumn{2}{c}{\textbf{Outdoor}}\\
\cline{3-6}
&   
&\textbf{NL}    &\textbf{LL}
&\textbf{NL}    &\textbf{LL}\\
\Xhline{0.7pt}
\multirow{2}{*}{EF-ANN} &  ball &  97.13\% & 93.63\%  & 97.75\%  &97.63\% \\
                       & human & 97.25\%  &97.75\%  & 98.25\%  &97.50\%   \\
\multirow{2}{*}{EST-ANN} &  ball &  97.75\% & 97.25\%  & 98.00\%  &97.88\% \\
                       & human & 97.75\%  &98.13\%  & 97.75\%  &97.63\%   \\
\multirow{2}{*}{EF-SNN} &  ball &  96.68\% & 89.75\%  & 97.63\%  &97.75\% \\
                       & human & 97.75\%  &94.63\%  & 96.75\%  &97.13\%   \\
\multirow{2}{*}{Ours} &  ball &  98.00\% & 97.25\%  & 98.13\%  &97.88\% \\
                       & human & 98.25\%  &97.50\%  & 97.25\%  &97.63\%   \\
\Xhline{1pt}
\end{tabular}
\label{Table:4}
\end{center}
\end{table}

\subsubsection{Comparison Results of Key-Event-Point Module}
\label{sec:kep module}

In this section, we analyze the effect of the proposed Key-Event-Point (KEP) module. 
we visualize the event streams before and after applying the KEP module using the event frame method, as shown in Fig.~\ref{fig:5}. 
The visualizations show that our module retains the object's clear edge features with fewer events and effectively removes background noise. 
Next, we calculate the dodging success rate and the average number of events in the event streams both before and after implementing the KEP module. 
The results, shown in Table~\ref{Table:5}, indicate that we achieved a success rate similar to the raw events using only 29.9\% of the ball events and 17.49\% of the human events.

\begin{table}[htbp]
\vspace{-1.0em}
\begin{center}
\caption{
Comparison Results of Key-Event-Point Module.
}
\vspace{-1.0em}
\setlength{\extrarowheight}{3pt}
\setlength{\tabcolsep}{0.02\linewidth}
\begin{tabular}{cccccccp{1cm}}
\Xhline{1pt}
\multirow{2}{*}{\textbf{Module}}
&\multirow{2}{*}{\textbf{Object}}
&\multicolumn{2}{c}{\textbf{Indoor}}
&\multicolumn{2}{c}{\textbf{Outdoor}}
&\multirow{2}{*}{\textbf{\makecell[c]{Average \\Points}}}\\
\cline{3-6}
&   
&\textbf{NL}    &\textbf{LL}
&\textbf{NL}    &\textbf{LL}\\
\Xhline{0.7pt}
\multirow{2}{*}{w/o KEP} &  ball &  98.00\% & 97.25\%  & 98.13\%  &97.88\% &1765\\
                       & human & 98.25\%  &97.50\%  & 97.25\%  &97.63\%   &3345\\
\multirow{2}{*}{w/ KEP} &  ball &  97.83\% & 96.50\%  & 98.13\%  &97.75\% &529\\
                       & human & 98.00\%  &97.63\%  & 97.25\%  &97.50\%   &585\\
\Xhline{1pt}
\end{tabular}
\label{Table:5}
\end{center}
\end{table}





\subsubsection{Power Consumption Comparison}
In this section, we evaluate the power consumption comparison of the following different methods on different devices: 1) EF-ANN and EST-ANN on NVIDIA Geforce RTX 4090 GPU, 2) EF-ANN and EST-ANN on NVIDIA Jetson NX Orin, 3) EF-SNN on Intel Kapoho Bay, and 3) Our method on Intel Kapoho Bay. 
We measure the static power, dynamic power, inference rate per second, and single inference power consumption. 
Dynamic power represents the device's power consumption during inference, and single inference power consumption is calculated by dividing the dynamic power by the inference rate per second. 
We use tools including nvidia-smi, energy monitoring, and the power consumption meter to monitor the average power of NVIDIA Geforce RTX GPU, NVIDIA Jetson Orin NX, and Kapoho Bay respectively. 
For the power measurement of NX Orin, we selected two modes: energy-saving mode (10 W) and high-performance mode (25 W). 
In the power consumption evaluation, the key event stream with a time window $\mathrm{W}=100$ is selected. 
The results are shown in Table~\ref{Table:6}.
The results show that the inference power consumption of EF-ANN is only 24.26\%, 21.4\%, and 20.67\% of that of EST-ANN in NVIDIA Geforce RTX GPU, NVIDIA Jetson NX Orin high-performance mode, and NX Orin energy-saving mode, respectively. 
This is because the method based on the event spike tensor uses additional convolution kernels, which improve robustness to different time windows, lighting conditions and objects but greatly increase the computational burden. 
In comparison, our method not only achieves similar robustness to EST-ANN but also significantly reduces inference power consumption. 
The inference power consumption of our method is only 0.95\% of that of EST-ANN on the GPU. 
Even in the energy-saving mode of NVIDIA Jetson NX Orin, our method consumes only 4.30\% of the power used by EST-ANN.
Additionally, compared with EF-SNN, our power consumption is reduced by 98.36\% on the same neuromorphic processor.

\begin{table*}[htbp]
\vspace{-1.0em}
\begin{center}
\caption{
Power consumption comparison of different methods on different devices.
}
\vspace{-1.0em}
\setlength{\extrarowheight}{3pt}
\setlength{\tabcolsep}{0.035\linewidth}
\begin{tabular}{ccccccp{1cm}}
\Xhline{1pt}
\textbf{Device}
&\textbf{Scheme}
&\textbf{Idle/mW}
&\textbf{Dynamic Consumer/mW}
&\textbf{Inf/s}
&\textbf{mJ/Inf}\\
\Xhline{0.7pt}
NVIDIA Geforce RTX 4090       &EF-ANN   &21000  &43000 &1179 &36.47\\
NVIDIA Geforce RTX 4090       &EST-ANN   &21000  &43000 &286 &150.35\\
NVIDIA Jetson Orin NX (25W)       &EF-ANN   &6123  &3971 &369 &10.76\\
NVIDIA Jetson Orin NX (25W)       &EST-ANN   &6119  &4157 &83 &50.08\\
NVIDIA Jetson Orin NX (10W)       &EF-ANN   &5944   &1630 &244 &6.68\\
NVIDIA Jetson Orin NX (10W)       &EST-ANN   &5942   &1906 &59 &32.31\\
Intel Loihi Kapoho Bay  &EF-SNN   &1865   &170 &2 &85 \\
Intel Loihi Kapoho Bay  &Ours   &1865   &170 &122 &1.39 \\
\Xhline{1pt}
\end{tabular}
\label{Table:6}
\end{center}
\end{table*}

\begin{figure*}[htbp]
\vspace{-0mm}
\centering
\subfigure[]{\label{fig:8a}\includegraphics[width=0.23\textwidth]{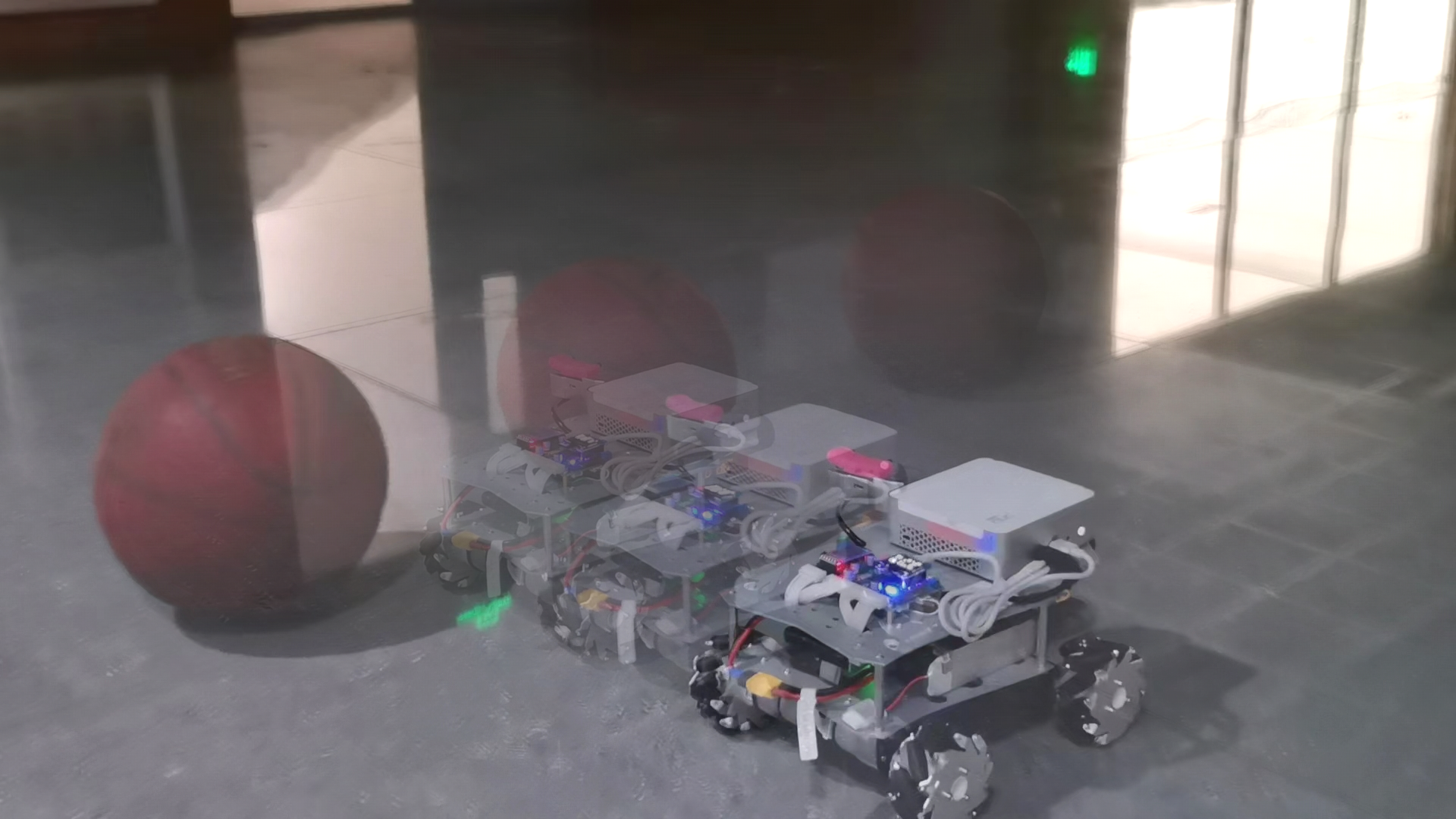}}
\subfigure[]{\label{fig:8b}\includegraphics[width=0.23\textwidth]{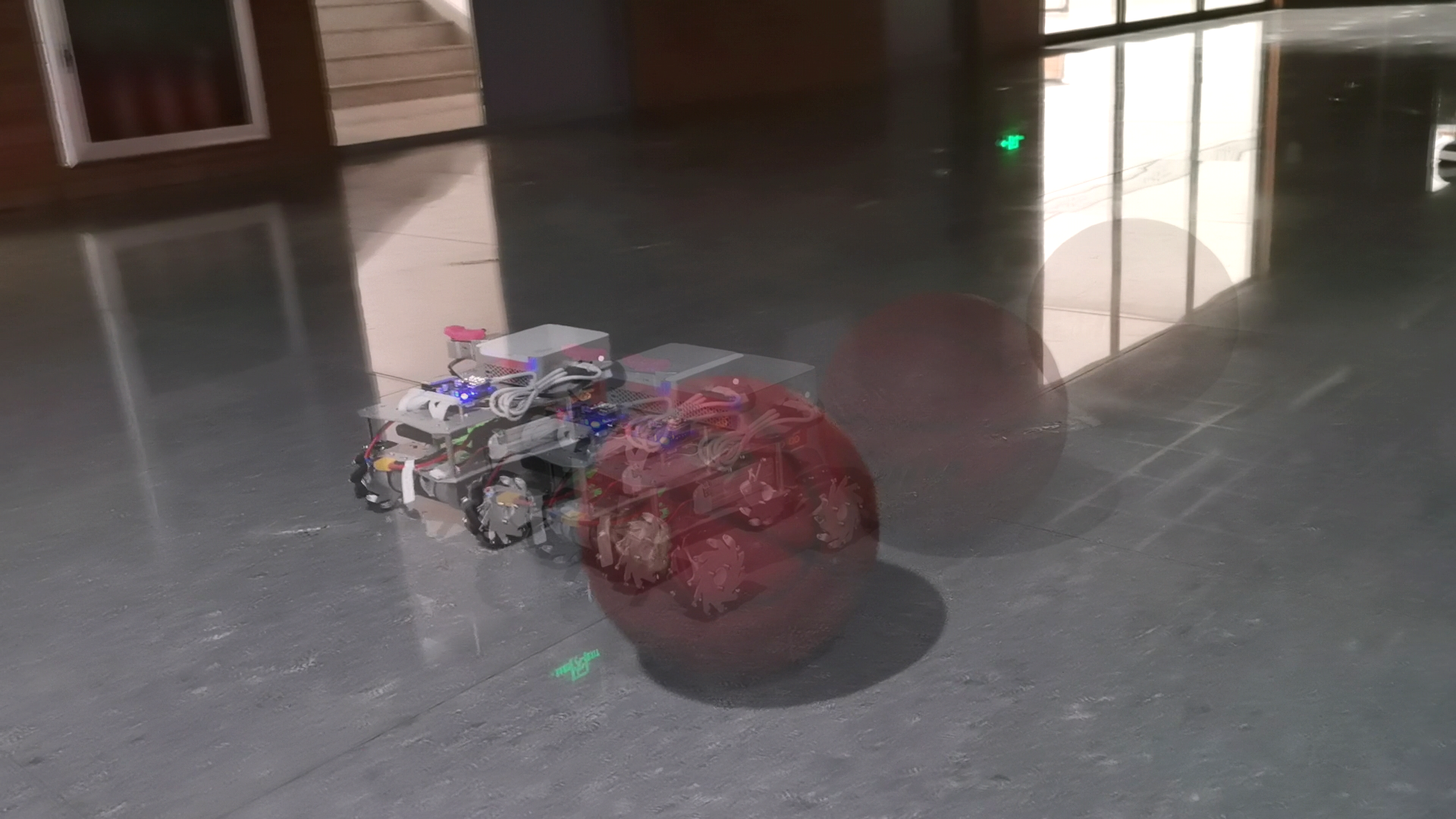}}
\subfigure[]{\label{fig:8c}\includegraphics[width=0.23\textwidth]{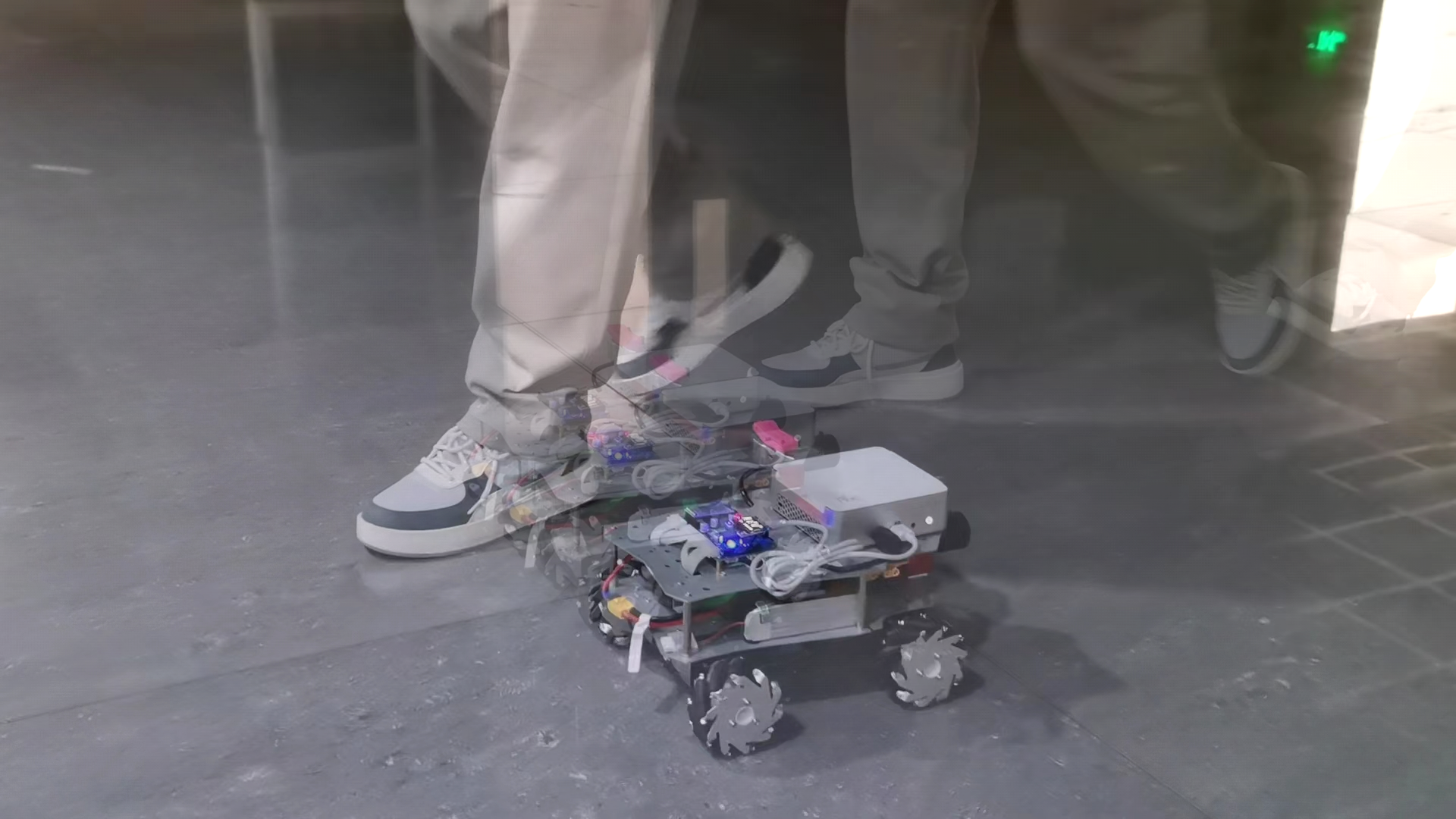}}
\subfigure[]{\label{fig:8d}\includegraphics[width=0.23\textwidth]{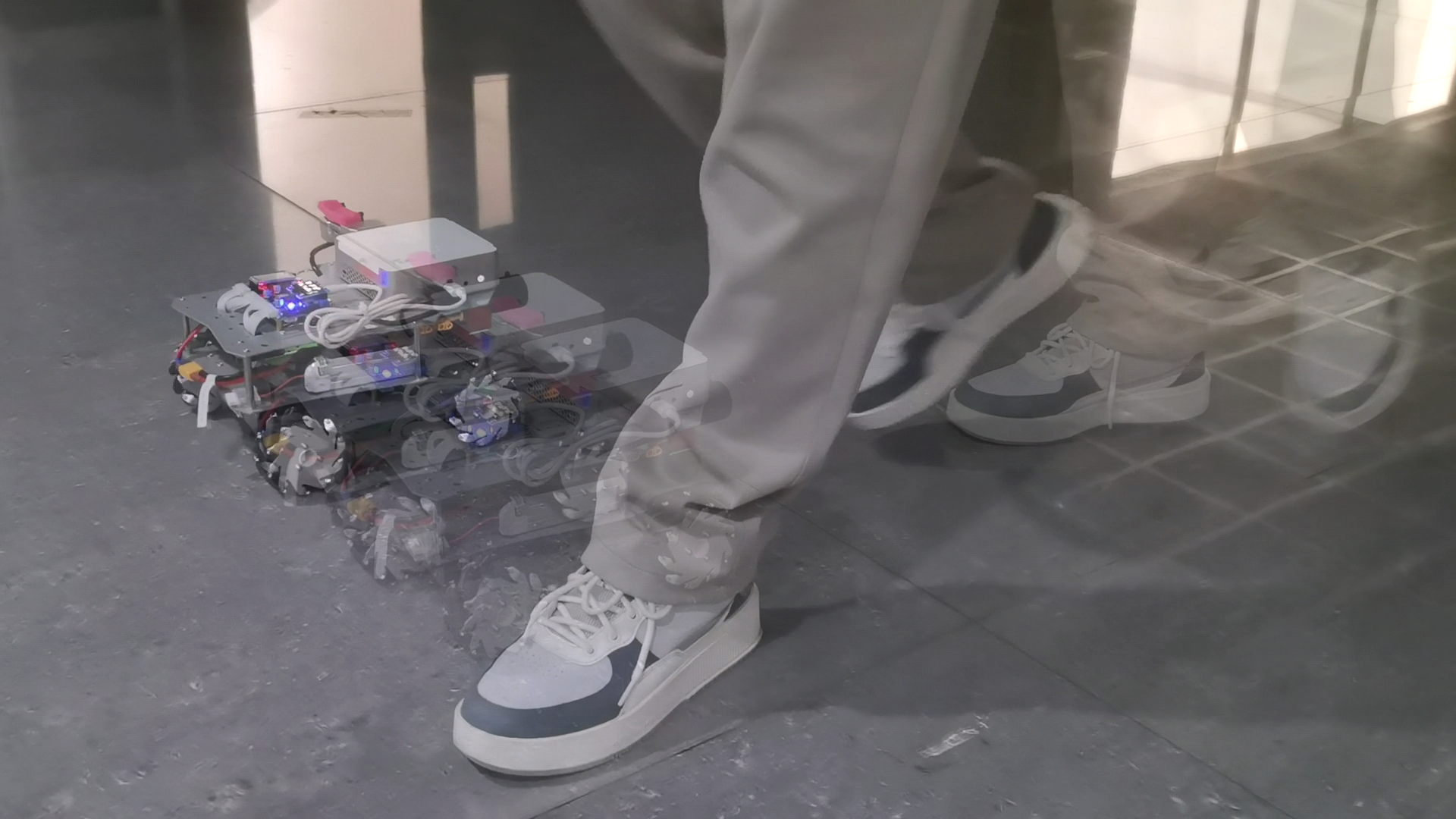}}
\vspace{-1em}
\caption{Physical experiments. 
We conduct a physical system verification in a normal indoor light environment. 
We position balls and humans to approach the mobile robot from the left and right directions. 
And the neuromorphic robot can dodge Successfully.}
\label{fig:8}
\vspace{-1em}
\end{figure*}
\subsection{Physical System Verification}
To verify the feasibility of the fully asynchronous neuromorphic paradigm we proposed for the mobile robot described in Section \ref{neu-robot}, we conducted a physical system verification in a normal indoor light environment. 
The mobile robot detects dynamic objects using the DAVIS 346. The host computer processes the raw event stream through the proposed KEP module, extracting the key event stream and converting it into an address sequence. The events are then asynchronously transmitted to the SNN based on this address sequence using embedded lakemont (x86) processors. After the SNN completes asynchronous processing, the host computer receives the output spike train and decodes it into the dodging action. Finally, the host computer sends the action command to the control board to execute the dodging maneuver.
The dodging effect is shown in Fig.~\ref{fig:8}.
Our system can successfully dodge both a person and a ball in time, effectively proving the feasibility of our method.

\section{Conclusion}
\label{sec:conclusions}

This article proposes a fully asynchronous neuromorphic perception paradigm and applies it to solve sequential tasks on real mobile robot for the first time. 
Our paradigm integrates event cameras, spiking neural networks, and neuromorphic processors to achieve low-power, robust perception for mobile robots dodging. 
To verify the superiority of the proposed perception paradigm, we conducted mobile robot dodging experiments using various methods with different time windows, lighting conditions and objects. 
We compared the proposed method with EF-ANN, EST-ANN, and EF-SNN. 
The results show that our method utilize the event streams Effectively and efficiently, making it more robust than EF-ANN and EF-SNN with different time windows and light conditions. 
Additionally, our method significantly reduces the computational burden compared to EST-ANN and EF-SNN. our method performs similarly to EST-ANN while consuming only 4.3\% of the power consumption of EST-ANN even in the NX Orin energy-saving mode.
Compared to Frame-SNN, our power consumption is reduced by 98.36\% on the same neuromorphic processor.

The proposed asynchronous neuromorphic paradigm in this article offers a more brain-like approach to perception. 
Although artificial intelligence and robots have achieved impressive results in various fields, there is still a significant gap in matching the ability of insects to perform various tasks in complex environments for extended periods. 
While many challenges remain in advancing from mobile robot dodging to mobile robot navigation, we believe that the proposed fully asynchronous neuromorphic perception paradigm, with its brain-like characteristics, will be an effective solution for future robots to perform difficult tasks in complex and changing environments.

\bibliographystyle{IEEEtran}
\bibliography{mybibfile}

\begin{IEEEbiography}[{\includegraphics[width=1in,height=1.25in,clip,keepaspectratio]{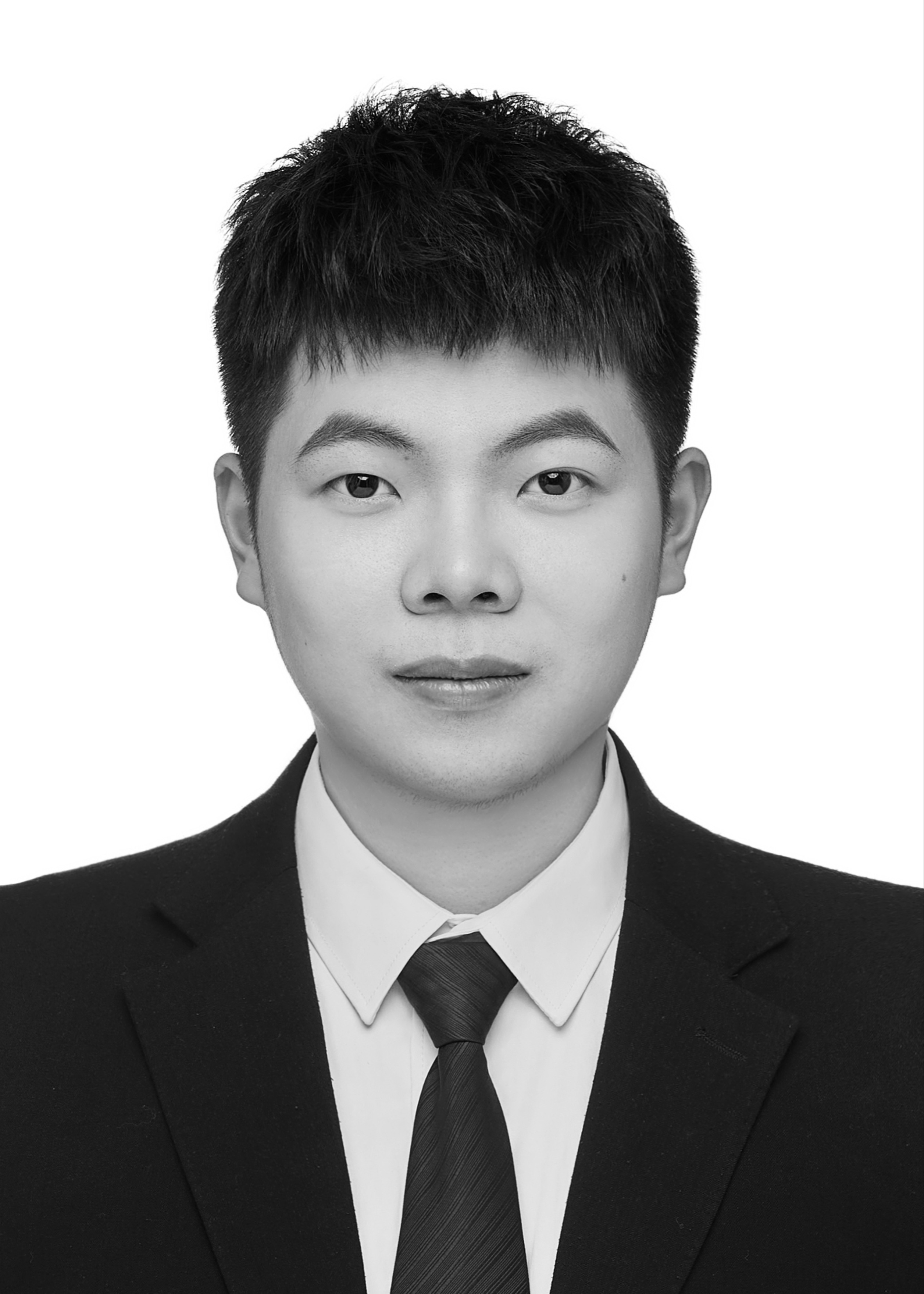}}]{Junjie Jiang}
received the B.S. degree in automation from Northeastern University at Qinhuangdao, Qinhuangdao, China, in 2020 and the M.S. degree in robot science and engineering from Northeastern University, Shenyang, China, in 2023. He is currently pursuing the Ph.D. degree in robot science and engineering with Northeastern University, Shenyang, China. His research interests include event-based vision, spiking neural network, robot visual navigation, and reinforcement learning.\end{IEEEbiography}
\vspace{-3em}

\begin{IEEEbiography}[{\includegraphics[width=1in,height=1.25in,clip,keepaspectratio]{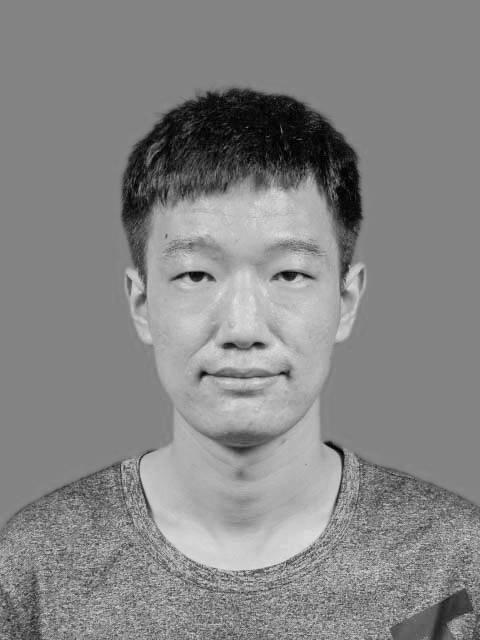}}]{Delei Kong}
(Graduate Student Member, IEEE) received the B.S. degree in automation from Henan Polytechnic University, Zhengzhou, China, in 2018, and the M.S. degree in control engineering from Northeastern University, Shenyang, China, in 2021.
From 2020 to 2021, he was an Algorithm Intern with SynSense Tech. Co. Ltd, Chengdu, China.
From 2021 to 2022, he was an R\&D Engineer (Advanced Vision) with the Machine Intelligence Laboratory, China Nanhu Academy of Electronics and Information Technology (CNAEIT), Jiaxing, China.
Since 2022, he has been a Ph.D. student in control science and engineering from Hunan University, Changsha, China. His research interests include event-based vision, robot visual navigation, and neuromorphic computing.\end{IEEEbiography}
\vspace{-3em}

\begin{IEEEbiography}[{\includegraphics[width=1in,height=1.25in,clip,keepaspectratio]{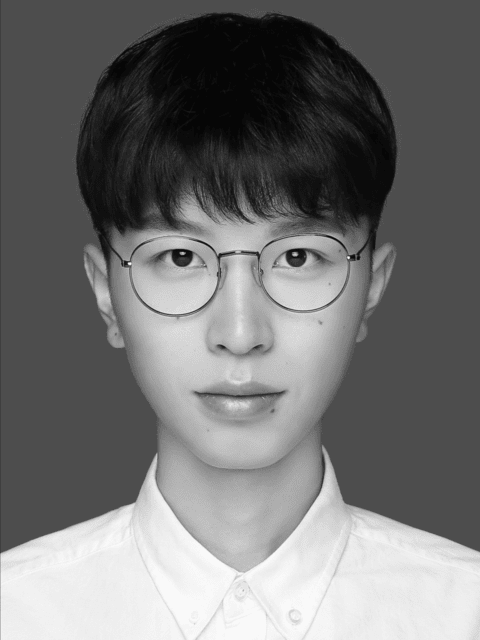}}]{Chenming Hu}
received the B.S. degree in robot engineering from Nanjing University of Information Science and Technology, Nanjing, China, in 2022. He is currently pursuing the M.S. degree in robot science and engineering with Northeastern University, Shenyang, China. His research interests include event-based vision, deep learning, and neuromorphic computing.\end{IEEEbiography}
\vspace{-3em}

\begin{IEEEbiography}[{\includegraphics[width=1in,height=1.25in,clip,keepaspectratio]{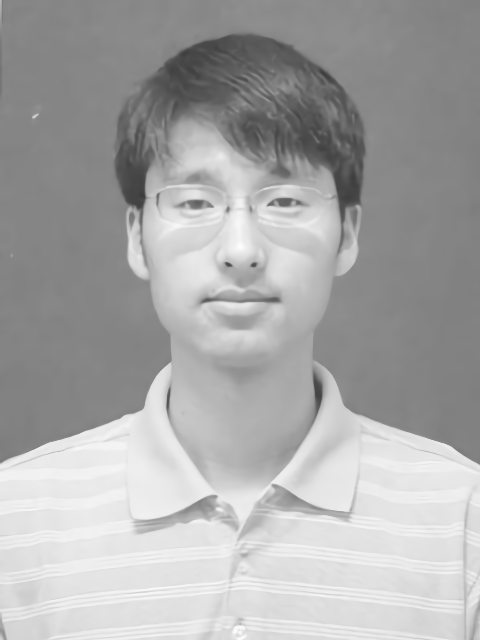}}]{Zheng Fang}
(Member, IEEE) received the B.S. degree in automation and the Ph.D. degree in pattern recognition and intelligent systems from Northeastern University, Shenyang, China, in 2002 and 2006, respectively. He was a Post-Doctoral Research Fellow at the Robotics Institute of Carnegie Mellon University (CMU), Pittsburgh, PA, USA, from 2013 to 2015. He is currently a full Professor with the Faculty of Robot Science and Engineering, Northeastern University. He has published over 80 papers in well-known journals or conferences in robotics and computer vision. His research interests include perception and autonomous navigation of various mobile robots, robot learning and neuromorphic computing.\end{IEEEbiography}

\end{document}